\documentclass[letterpaper,twocolumn,10pt]{article}
\usepackage{goodstuff}
\usepackage{usenix2019_v3}

\usepackage{tikz}
\usepackage{amsmath}
\usepackage{tightenitemize}

\usepackage{filecontents}

\begin{document}

\date{}

\title{\Large \bf Provenance-Based Interpretation of Multi-Agent Information Analysis}

\author{
{\rm Scott Friedman, Jeff Rye, David LaVergne, Dan Thomsen}\\
SIFT, LLC \\
\{friedman, rye, dlavergne, dthomsen\}@sift.net
\and
{\rm Matthew Allen and Kyle Tunis}\\
Raytheon BBN Technologies \\
\{matthew.allen, kyle.h.tunis\}@rtx.com
} 

\maketitle

\begin{abstract}
Analytic software tools and workflows are increasing in capability, complexity, number, and scale, and the integrity of our workflows is as important as ever.
Specifically, we must be able to inspect the process of analytic workflows to
assess (1) confidence of the conclusions, (2) risks and biases of the operations involved, (3) sensitivity of the conclusions to sources and agents, (4) impact and pertinence of various sources and agents, and (5) diversity of the sources that
support the conclusions.
We present an approach that tracks agents' provenance with PROV-O in conjunction with agents' appraisals and evidence links (expressed in our novel DIVE ontology).
Together, PROV-O and DIVE enable dynamic propagation of confidence and counter-factual refutation to improve human-machine trust and analytic integrity.
We demonstrate representative software developed for user interaction with that provenance, and discuss key needs for organizations adopting such approaches.
We demonstrate all of these assessments in a multi-agent analysis scenario, using an interactive web-based information validation UI.

\end{abstract}

\section{Introduction}
\label{sec:intro}

Data-intensive workflows--- ranging from intelligence analysis to journalism to computational biology--- increasingly rely on advanced software technology to facilitate analysis.
Advanced software may expedite results and extend analytic capability, but often with increased complexity, increased technological risk, or loss of human interpretability.

Despite the increased complexity, our core principles and metrics for integrity, quality \cite{icd_203}, and pertinence \cite{icd_206} remain as important as ever, perhaps even \emph{moreso} than ever.
Aside from the integrity of the result we must ensure that the human-machine collaboration \emph{process} is conducted with analytic rigor \cite{zelik2010measuring} and enables human understanding and trust of the constituent operations \cite{klein2007data}.

This paper presents an approach to dynamically validate and explore information produced by automated software agents, inspired in part by recent work on provenance-based label propagation (e.g., \cite{han2020unicorn}) and decision provenance \cite{singh2018decision}.
We prevent a novel DIVE (Dynamic Information Validation and Explanation) ontology, a minimal extension of the PROV ontology \cite{lebo2013prov} for expressing agents' appraisals, assumptions, and evidence over the data.
We build on graph propagation and and truth-maintenance algorithms \cite{forbus1993building,de1986assumption}, and we extend these with novel classes and semantic constraints to represent the derivation and rationale for conclusions, the appraisals of various agents on those conclusions, and the propagation of confidence from sources to conclusions.

We apply our approach in a simplified intelligence analysis domain, where outcomes are derived along multiple paths by multiple autonomous agents.
We focus primarily on the inter-agent flow rather than the inner workings (e.g., inference engines and machine learning models) of individual agents.
We are interested in assessing the integrity of these flows and of the outcomes they support, modulo the confidence, assumptions, diversity, and sensitivity of upstream sources.

We continue with a review of provenance-tracking and truth-maintenance algorithms.
\secref{approach} describes our knowledge representation and reasoning approach using provenance as a platform for explanation.
\secref{experiments} presents empirical results of our system generating explanations, and we review the results and outline future work in \secref{conclusions}.

\hide{
\begin{figure}
  \begin{center}
  \includegraphics[width=\linewidth]{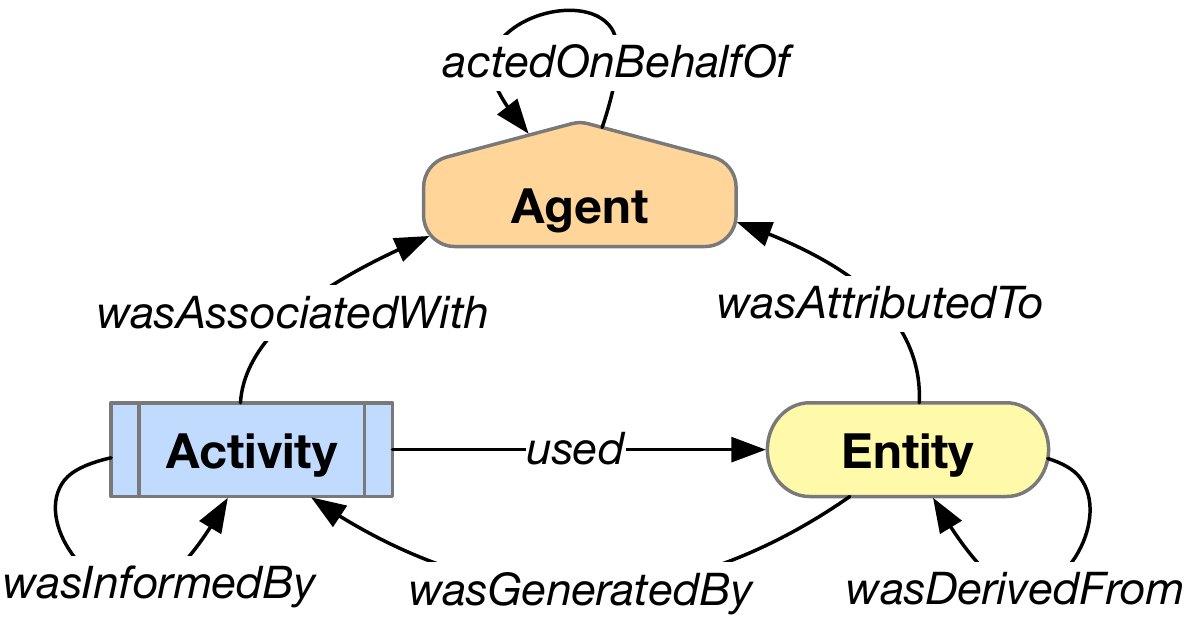}
  \caption{PROV ontology subset used in our approach.}
  \label{fig:prov}
  \end{center}
  \vspace{-.2in}
\end{figure}
}

\subsection{Provenance-Tracking}
\label{sec:prov}

Our technical approach extends the PROV-O ontology \cite{lebo2013prov}, which expresses provenance entities and relationships using the OWL2 Web Ontology Language.
The PROV Data Model includes the following three primary classes of elements to express provenance:
\begin{enumerate}
  \item \textbf{Entities} are real or hypothetical things with some fixed aspects in physical or conceptual space.  These may be assertions, documents, databases, inferences, etc.
  \item \textbf{Activities} occur over a period of time, processing and/or generating entities.  These may be inference actions, judgment actions, planned (not yet performed) actions, etc.
  \item \textbf{Agents} are responsible for performing activities or generating entities.  These may be humans, web services, machine learning modules, etc.
\end{enumerate}

Systems that utilize PROV can represent long inferential chains, formally linking conclusions (e.g., a downstream assertion) through generative activities (e.g., inference operations) and antecedents, to source entities and assumptions.
This comprises a directed network of provenance that we can traverse in either direction to answer questions of foundations, derivations, and impact.

\subsection{Truth-Maintenance Systems}
\label{sec:tms}

Truth-Maintenance Systems (TMSs) \cite{forbus1993building,friedman2018csj} explicitly store entities alongside \emph{justifications} that link antecedent entities (analogous to PROV entities) with consequent entities.
This explicitly encodes the rationale for each entity, so --- similar to the PROV ontology --- we can use a TMS to explore foundations, derivations, and impact.

TMSs track \emph{environments} as sets of elements that sufficiently justify an entity in its upstream lineage.
If the lineage changes (e.g., due to a new derivation of an entity), the TMS recomputes the affected environments.
Environments allow TMSs to efficiently recognize contradictions, retrieve logical rationale, and identify upstream assumptions \cite{de1986assumption}.
TMSs operate alongside inference engines to record the lineage and logical conditions for believing various assertions; they do not themselves generate inferences or derive entities.
Our approach utilizes TMS-like environments to efficiently refute information, propagate confidence, and visualize impact.

\section{Approach}
\label{sec:approach}


\begin{figure}
  \begin{center}
  \includegraphics[width=\linewidth]{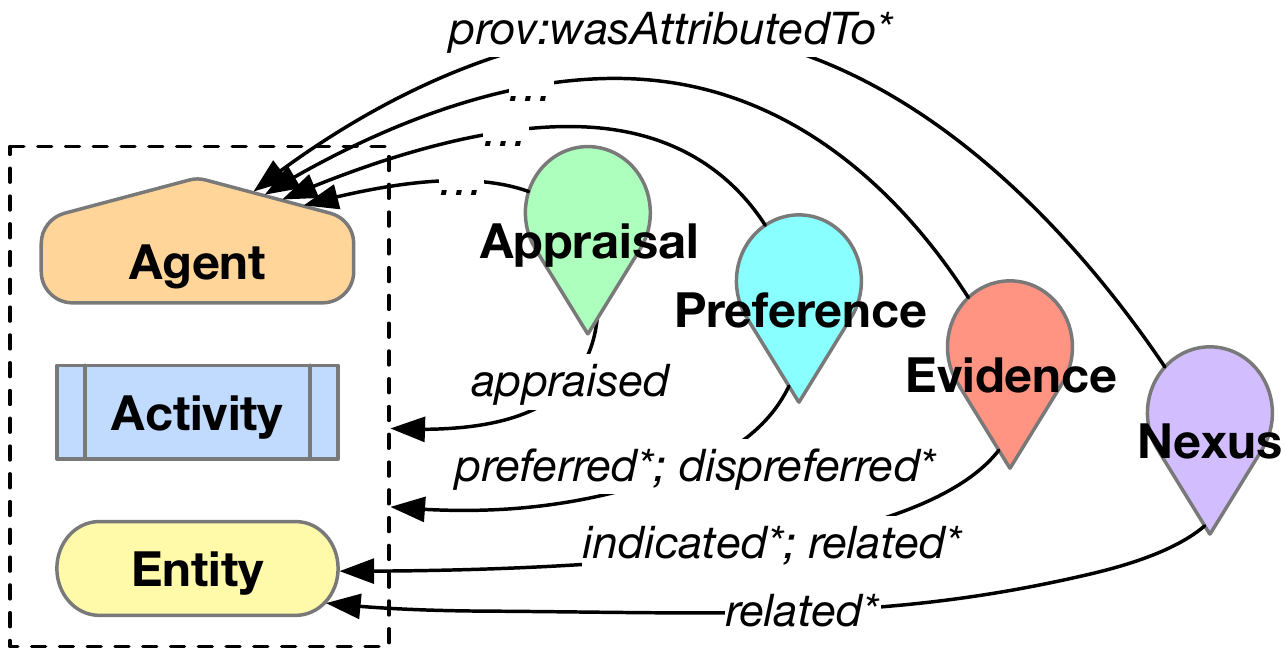}
  \caption{DIVE ontology classes and relations, using some elements of the PROV ontology.}
  \label{fig:dive}
  \end{center}
  \vspace{-.2in}
\end{figure}

\subsection{Semantic Extensions and Constraints}
\label{sec:kr}

Our novel DIVE ontology, illustrated in \figref{dive} is a minimal extension to the PROV ontology that introduces four additional classes and relations that relate to the PROV ontology.
The edges to the dotted box in \figref{dive} indicate that the edge (e.g., \textbf{appraised}) can target any class contained within.

\begin{enumerate}

  \item An \textbf{Appraisal} is a human or machine agent's judgment about an activity, entity, or other agent.
    There is at most one appraisal for any appraising (\textbf{prov:wasAttributedTo}) agent and \textbf{appraised} element.  Appraisals have attributes to describe confidence and likelihood judgments using ICD 203 metrics \cite{icd_203}.  Different agents may appraise an element differently.

  \item \textbf{Evidence} is agent judgment about diagnosticity of one entity on another.  This includes evidence and counter-evidence.
    Evidence is \emph{directed} from \textbf{related} entities to \textbf{indicated} entities.
    As with appraisals, different agents may express conflicting evidence.

  \item A \textbf{Preference} is an agent's relative judgment about the relative quality or confidence of one entity, activity, or agent over another.  Preferences are not absolute judgments, but relative \emph{ceteris paribus} judgments.  This means a machine agent may express a preference of one of its inferences over another, \emph{all else being equal}.

  \item A \textbf{Nexus} is an agent's judgment over a \emph{set} of entities to qualitatively or numerically express mutual coherence (e.g., high joint likelihood) or conflict (e.g., low joint likelihood), using ICD 203 metrics \cite{icd_203}.

\end{enumerate}

These DIVE classes express human and machine attributions about the quality of the entities, activities, and agents in the PROV-O record.
Consequently, DIVE is expressed at the meta-level of PROV.

Taken together, PROV is a network of information generation and information flow, and DIVE expresses agent judgments about the information and flow.
These judgments extend the PROV network and flow through the network to facilitate downstream quality judgments and interpretation.

\begin{figure*}[htb]
  \begin{center}
    \frame{
  \includegraphics[width=\textwidth]{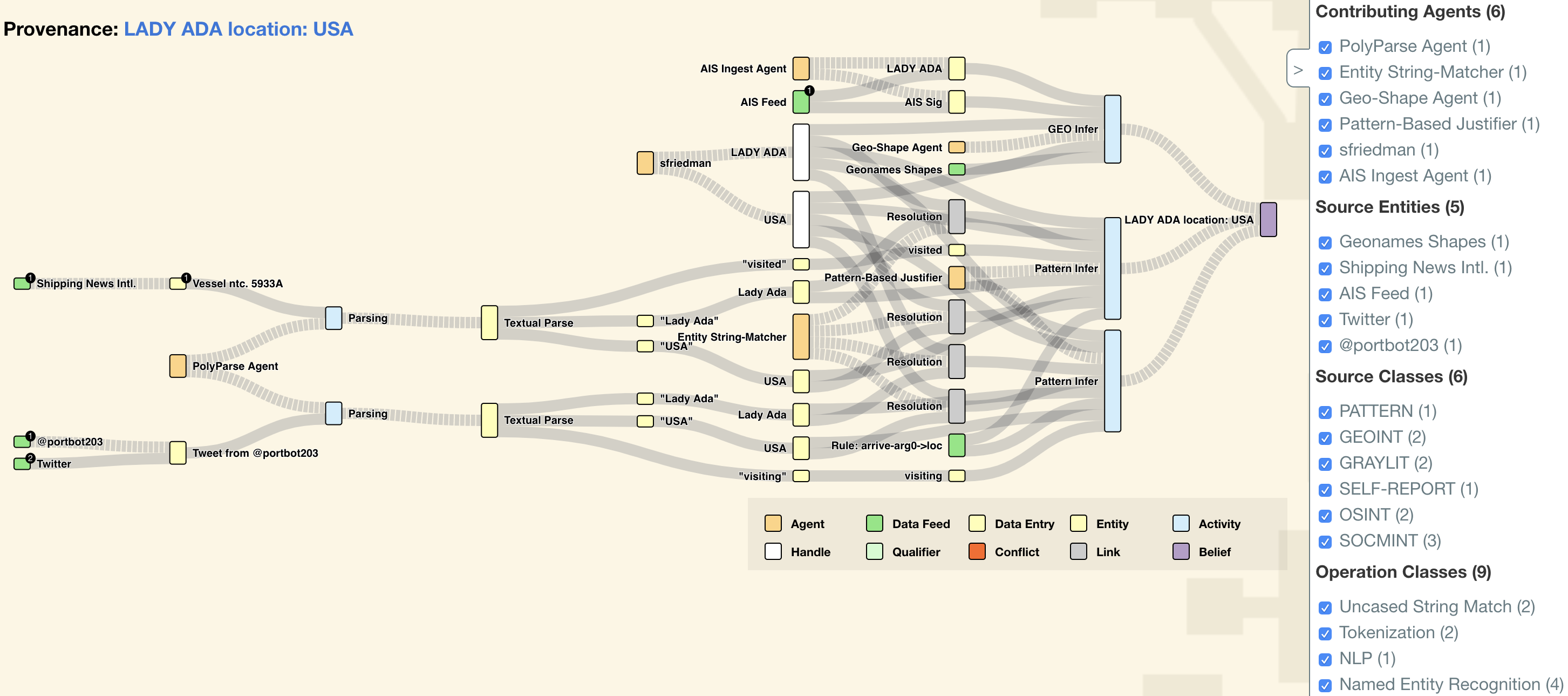}
  }
  \caption{Screenshot from a live provenance visualization, plotting the full human-machine provenance for the assertion that a vessel ``Lady Ada'' was at one point located in the USA.}
  \label{fig:ss-prov}
  \end{center}
  \vspace{-.3in}
\end{figure*}

\subsection{Provenance Retrieval and TMS Structure}
\label{sec:retrieval}

We implemented a multi-agent information analysis platform using JanusGraph\footnote{https://janusgraph.org/} as a shared knowledge graph for workflow and provenance-tracking.
JanusGraph uses Apache TinkerPop\footnote{http://tinkerpop.apache.org/} graph computing framework, so our system runs a TinkerPop-based PROV-O graph traversal to retrieve the full upstream provenance for any set of target assertions.

After retrieving the upstream provenance elements, the system generates a TMS-like structure from the provenance: PROV activities act as TMS justifications that join the antecedent entities and agents with consequent entities.
This produces an AND/OR graph detailing the (one or more) derivations of all assertions in the upstream provenance structure.

The system then computes the TMS environments for each element to express the set of necessary and sufficient upstream PROV entities, agents, and activities supporting said element.

\subsection{Indexing for Dynamic Interpretation}
\label{sec:indexing}

Given a provenance graph to assess, our provenance system identifies and catalogs the following elements of the provenance structure:
\begin{itemize}
  \item \textbf{Agents}: actors in the analysis, defined in PROV-O.
  \item \textbf{Sources}: individual devices or informational resources from which information is derived, such as databases, websites, news agencies, human sources, and sensors.
  \item \textbf{Source classes}: categories of information, spanning potentially many assertions and information sources.
  \item \textbf{Operation classes}: categories of analytic activities, such as NLP and Pattern-Based Inference.
\end{itemize}
Our system computes the TMS environments of all nodes in the provenance graph, relative to the above catalogs of elements.
This means that each activity and entity in the graph is indexed by the agents, sources, source classes, and operation classes that were necessary or sufficient to execute the activity or derive the assertion, respectively.

\section{Propagation \& Visualization Results}
\label{sec:experiments}

We demonstrate our approach in a simplified fictional intelligence analysis exercise.
Our objective in this work is to validate information that flows across agents, opposed to exploring the within-agent machine learning pipelines.
Consequently, agent actions (e.g., NLP) are represented as atomic activities rather than as massive sub-networks.

We implemented our provenance browser as a graphical display within a larger web-based platform for human-machine collaborative intelligence analysis.
At any time in their analysis, the user may select one or more entities from a diagram or listing and opt to view their provenance.

A webservice traverses the knowledge graph to retrieve the full provenance for the desired entities, and all relevant Appraisals therein, and sends it to the client.
The client's provenance visualizer uses JavaScript to implement the retrieval, refutation, and propagation algorithms described above, operating over the PROV and DIVE representations.

\figref{ss-prov} shows a screenshot of the provenance for the assertion that a cargo ship ``Lady Ada'' was located in the USA (\figref{ss-prov}, rightmost node), along three different derivation paths, immediately to its left: (1) a GEOINT path using the vessel's AIS transponder (top); (2) a NLP and pattern-based inference path from a fictitious ``Shipping News International'' source; and (3) a similar NLP path from a fictitious Twitter post.
On the right of \figref{ss-prov} is a sidebar cataloging the analytic factors detailed in \secref{indexing}.

Using this example, we describe three general provenance-based exploration strategies to help human operators dynamically isolate, refute, and critically analyze the sensitivity and confidence of complex human-machine analyses.

\begin{figure*}[h!]
  \begin{center}
  \includegraphics[width=\textwidth]{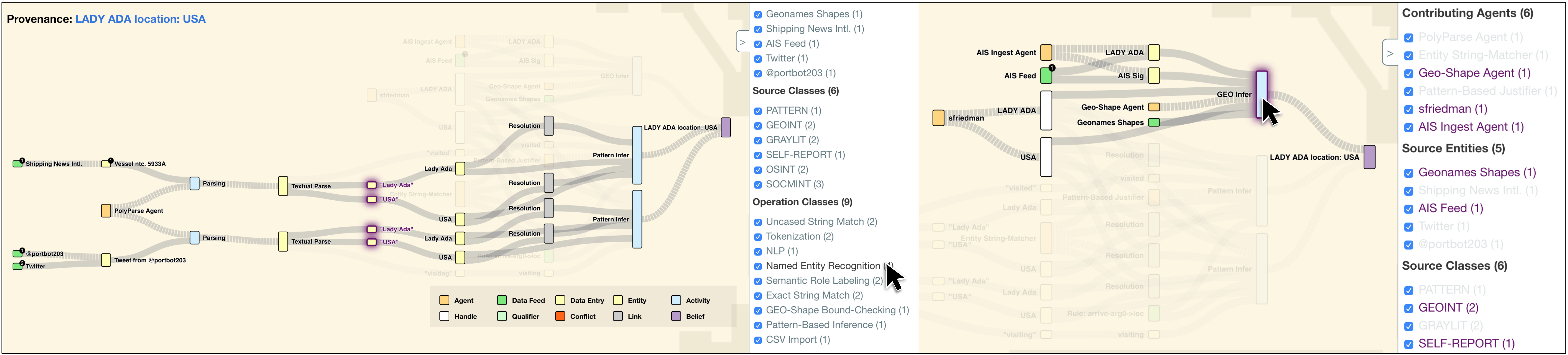}
  \caption{Isolating the analytic contribution of Named Entity Recognition (left) and a single GEOINT inference (right).}
  \label{fig:highlight}
  \end{center}
  \vspace{-.1in}
\end{figure*}

\begin{figure*}[h!]
  \begin{center}
  \includegraphics[width=\textwidth]{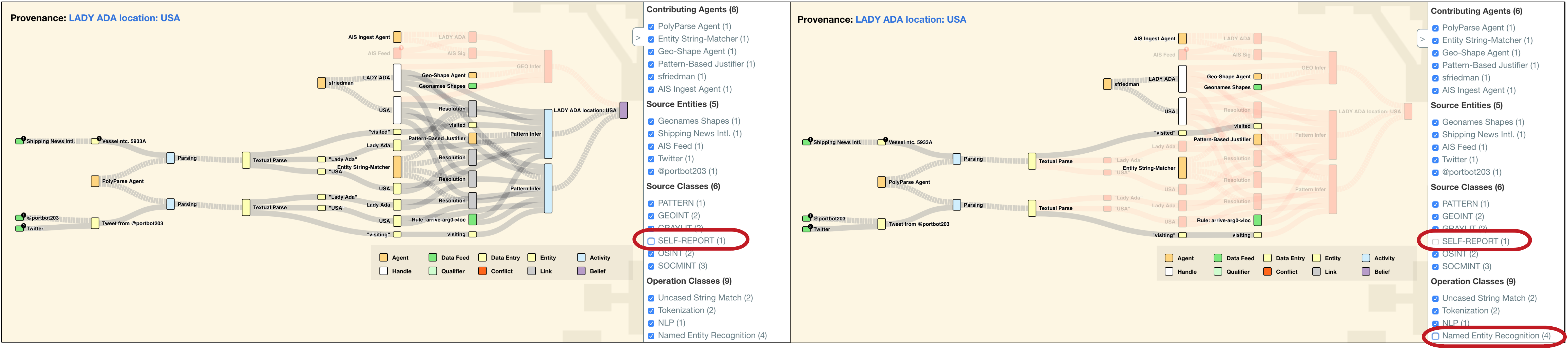}
  \caption{Refuting SELF-REPORT (left) and \emph{both} NER and Self-Reported data, which ultimately refutes the assertion (right).}
  \label{fig:refutation}
  \end{center}
  \vspace{-.2in}
\end{figure*}

\subsection{Isolating Flows with Environments}
\label{sec:hilight}

Raw derivation trees often contain too much data to assess in their entirety, and the volume of \textbf{Activities} increases the breadth and depth of the graph.
To manage complexity, we can isolate flows to inspect the contribution of an activity, agent, source, source class, or operation class.

Hovering over any cataloged element in the right-hand index isolates the graph to the contribution of that element alone.
\figref{highlight} (left) shows the effect of hovering over Named Entity Recognition (NER): four named entities (the ``Lady Ada'' and ``USA'' from two different sources) are highlighted in the graph, and all elements are de-emphasized except (1) upstream elements in the NER entities' environments and (2) downstream elements that have NER within their environments.
This facilitates \emph{source pertinence} judgments \cite{icd_206} and also \emph{impact analysis} to see the contribution and input flows of any operation,  or class thereof.

Similarly, hovering over any element in the graph displays the upstream and downstream elements, using the TMS environments in a similar fashion.
\figref{highlight} (right) shows the effect of hovering over the ``Geo Infer'' activity in the graph: the relevant upstream and downstream flows are selected via their environments, and the right-hand catalogs highlight only the sources and classes involved upstream.
This facilitates \emph{drill-down analysis} to isolate individual elements and understand complex derivations in relevant subsets.

\subsection{Refutation with Environments}
\label{sec:refutation}

We use TMS environments to \emph{refute} elements or cataloged classes.
\figref{refutation} (left) illustrates the effect of disabling ``SELF-REPORT'' data, since the vessel reports its own AIS signals.
This disables the Geo-Infer derivation and other elements that necessarily depend on disabled elements (i.e., where all TMS environments are [partially] refuted).
One of the derivations are partially disabled, but the other two (Pattern Inferences) remain.
If we \emph{also} disable ``Named Entity Recognition,'' this temporarily disables the NER-tagged elements as well as the target assertion in \figref{refutation} (right).

This refutation capability facilitates \emph{sensitivity analysis} \cite{zelik2010measuring, icd_203} to any source, activity, operation class, or source class, since it allows us to counter-factually see the analysis without that contribution.
This also helps us perform a \emph{risk assessment}, e.g., to see how much our analysis depends on potentially-risky machine operations.

\begin{figure*}[ht!]
  \begin{center}
  \includegraphics[width=\textwidth]{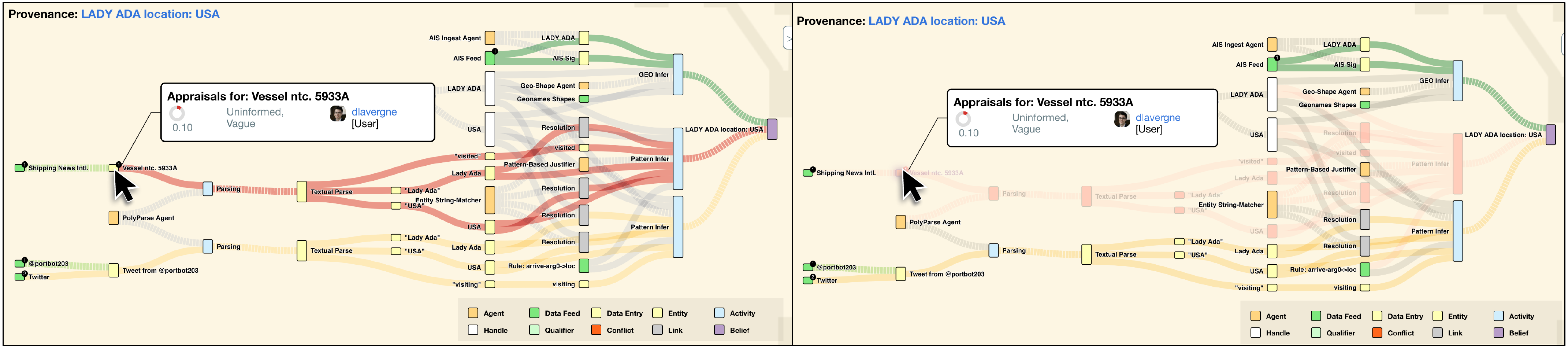}
  \caption{Propagating all attributions through the provenance (left) and individually refuting a low-confidence document--- and all of its individual low-confidence contribution--- from the analysis (right).}
  \label{fig:confidence}
  \end{center}
  \vspace{-.2in}
\end{figure*}

\subsection{Propagating Appraisals}
\label{sec:propagation}

If a human or machine agent creates a DIVE Appraisal, Preference, or Nexus, we can propagate the numerical data through the network.
\figref{confidence} (left) shows a very low-confidence Appraisal from a human agent about a document (0.1 on the ICD-203 scale \cite{icd_203}), and a forward-propagation of confidence scores throughout the graph, where red indicates low confidence and green indicates high confidence.
Notice that the ``Shipping News International'' organization confidence starts relatively high, but the low-confidence document authored by that organization immediately dominates it downstream in the remainder of the analysis.
We see that the target assertion (at right) is a junction of a high-, low-, and moderate-confidence derivations.
If we refute the low-confidence document (\figref{confidence}, right), we excise the low-confidence elements from our analysis and our assertion still holds.

This helps inform multiple dimensions analytic rigor \cite{zelik2010measuring} such as \emph{information validation} (i.e., verifying the information contribution from relevant sources), \emph{information synthesis} (i.e., incorporating data and considering the diversity of inferences) and elements of \emph{specialist collaboration} (i.e., using experts' consultation on relevant topics or sources).

We allow the user to select among multiple policies for propagating confidence, such as \emph{minimum}, \emph{maximum}, and \emph{average} to handle \emph{and}/\emph{or} junctions of confidence, but more sophisticated Bayesian approaches (e.g., \cite{kuter2007sunny}) can be integrated to express more complicated confidence propagation strategies, e.g., where the confidence of a multiply-derived assertion is higher-confidence than any of its constituent sources due to the added confidence of diverse corroboration.

\section{Related Work}
\label{sec:related}

Previous work summarizes information flows to detect persistent security threats in real time \cite{han2020unicorn,newsome2005dynamic} by propagating labels through newtork flows.
Other work uses cloud-based multi-agent provenance across source domains \cite{gehani2010mendel} and as a service (e.g., \cite{pasquier2016information}), which we believe is compatible with our approach to helping users understand how information was derived, or how decisions are made in a complex pipeline \cite{singh2018decision}.
Other systems track lineage in large databases \cite{benjelloun2008databases} and in multi-agent analysis \cite{toniolo2015supporting} but do not have the per-agent expressivity and activity-level refutation of our DIVE ontology.
Label propagation approaches to have been used for intent recognition with support for refutation \cite{primrose2018}, but this does not use formal provenance notation, evidence annotations, or multi-agent appraisal.

\section{Conclusions and Future Work}
\label{sec:conclusions}

This paper presented an integrated approach for using provenance to dynamically explore complex multi-agent analyses.
Our integrated approach includes the following technical contributions: (1) the DIVE ontology to allow PROV agents to express evidence, appraisals, preferences, and nexuses; (2) adaptations of TMS environments into the provenance framework; and (3) environment-level indexing by sources, agents, and operations; and (4) visualization techniques to support appraisal, isolation, refutation, and propagation of elements and agent insights.
We identified specific analytic integrity directives \cite{icd_203, icd_206} and dimensions of analytic rigor \cite{zelik2010measuring} facilitated by our approach, ultimately improving the ability for people to reason intuitively and efficiently about complex human-machine analyses.

Our approach is designed to visualize and validate higher-level inter-agent flow across data sources.
For data-intensive scientific computing and massive machine learning pipelines, our graph propagation and TMS environment structure would still apply, but our full visualization approach would not be informative without filtering, e.g., to prioritize and display only elements that most impact confidence.

One critical assumption we make in this work is that each software agent in the human-machine team logs its provenance soundly, completely, and at the right granularity.
Intuitively, if machine agents violate this assumption, our refutation and confidence models become unusable, or worse, misleading.
This presents a governance issue: agents can only be admitted into the analytic framework if their provenance-logging satisfies these assumptions.


Empirically validating our approach with a user study is an important next step.
This will help us characterize the effect of these provenance-based analyses on the rigor of the analytic process and the user's situation awareness.

Finally, the complex provenance graphs displayed and manipulated in our approach are not necessarily \emph{explanations}, but they contain structure that can support explanations of users' \emph{how}, \emph{why}, and \emph{what-if} questions.
Consequently, we see value in using the approaches presented here in conjunction with additional reasoning for machine Q\&A about complex workflows with dynamic provenance displays.

\section*{Acknowledgments}

This research was supported by the Air Force Research Laboratory.
Distribution A.  Approved for public release; distribution unlimited.
88ABW-2020-1387; Cleared 15 Apr 2020.

\bibliographystyle{plain}
\bibliography{proj7,sensemaking,provenance}

\begin{thebibliography}{10}

\bibitem{benjelloun2008databases}
Omar Benjelloun, Anish~Das Sarma, Alon Halevy, Martin Theobald, and Jennifer
  Widom.
\newblock Databases with uncertainty and lineage.
\newblock {\em The VLDB Journal}, 17(2):243--264, 2008.

\bibitem{de1986assumption}
Johan De~Kleer.
\newblock {{An assumption-based TMS}}.
\newblock {\em Artificial intelligence}, 28(2):127--162, 1986.

\bibitem{forbus1993building}
Kenneth~D Forbus and Johan De~Kleer.
\newblock {\em Building problem solvers}, volume~1.
\newblock MIT press, 1993.

\bibitem{friedman2018csj}
Scott Friedman, Kenneth Forbus, and Bruce Sherin.
\newblock Representing, running, and revising mental models: A computational
  model.
\newblock {\em Cognitive Science}, 42(4):1110--1145, 2018.

\bibitem{gehani2010mendel}
Ashish Gehani and Minyoung Kim.
\newblock Mendel: Efficiently verifying the lineage of data modified in
  multiple trust domains.
\newblock In {\em Proceedings of the 19th ACM International Symposium on High
  Performance Distributed Computing}, pages 227--239, 2010.

\bibitem{primrose2018}
Robert Goldman, Scott Friedman, and Jeffrey Rye.
\newblock Plan recognition for network analysis: Preliminary report.
\newblock In {\em AAAI Workshop on Plan, Activity and Intent Recognition,
  February 2018}, 2018.

\bibitem{han2020unicorn}
Xueyuan Han, Thomas Pasquier, Adam Bates, James Mickens, and Margo Seltzer.
\newblock {{UNICORN: Runtime Provenance-Based Detector for Advanced Persistent
  Threats}}.
\newblock {\em arXiv preprint arXiv:2001.01525}, 2020.

\bibitem{klein2007data}
Gary Klein, Jennifer~K Phillips, Erica~L Rall, and Deborah~A Peluso.
\newblock A data-frame theory of sensemaking.
\newblock In {\em Expertise out of context: Proceedings of the sixth
  international conference on naturalistic decision making}, pages 113--155.
  New York, NY, USA: Lawrence Erlbaum, 2007.

\bibitem{kuter2007sunny}
Ugur Kuter and Jennifer Golbeck.
\newblock Sunny: A new algorithm for trust inference in social networks using
  probabilistic confidence models.
\newblock In {\em AAAI}, volume~7, pages 1377--1382, 2007.

\bibitem{lebo2013prov}
Timothy Lebo, Satya Sahoo, Deborah McGuinness, Khalid Belhajjame, James Cheney,
  David Corsar, Daniel Garijo, Stian Soiland-Reyes, Stephan Zednik, and Jun
  Zhao.
\newblock {{PROV-O: The PROV Ontology}}.
\newblock {\em W3C recommendation}, 30, 2013.

\bibitem{newsome2005dynamic}
James Newsome and Dawn~Xiaodong Song.
\newblock Dynamic taint analysis for automatic detection, analysis, and
  signature generation of exploits on commodity software.
\newblock In {\em NDSS}, volume~5, pages 3--4. Citeseer, 2005.

\bibitem{icd_206}
{Office of the Director of National Intelligence}.
\newblock Intelligence community directive 206: Sourcing requirements for
  disseminated analytic products, 2007.

\bibitem{icd_203}
{Office of the Director of National Intelligence}.
\newblock Intelligence community directive 203: Analytic standards, 2015.

\bibitem{pasquier2016information}
Thomas FJ-M Pasquier, Jatinder Singh, Jean Bacon, and David Eyers.
\newblock Information flow audit for {{PaaS}} clouds.
\newblock In {\em 2016 IEEE International Conference on Cloud Engineering
  (IC2E)}, pages 42--51. IEEE, 2016.

\bibitem{singh2018decision}
Jatinder Singh, Jennifer Cobbe, and Chris Norval.
\newblock Decision provenance: Harnessing data flow for accountable systems.
\newblock {\em IEEE Access}, 7:6562--6574, 2018.

\bibitem{toniolo2015supporting}
Alice Toniolo, Timothy Norman, Anthony Etuk, Federico Cerutti, Robin~Wentao
  Ouyang, Mani Srivastava, Nir Oren, Timothy Dropps, John~A Allen, and Paul
  Sullivan.
\newblock Supporting reasoning with different types of evidence in intelligence
  analysis.
\newblock {\em Proceedings of the 14th International Conference on Autonomous
  Agents and Multiagent Systems (AAMAS 2015)}, 2015.

\bibitem{zelik2010measuring}
Daniel~J Zelik, Emily~S Patterson, and David~D Woods.
\newblock Measuring attributes of rigor in information analysis.
\newblock {\em {{Macrocognition metrics and scenarios: Design and evaluation
  for real-world teams}}}, pages 65--83, 2010.

\end{thebibliography}

\end{document}